\documentclass[conference]{IEEEtran}

\usepackage{amsmath}
\usepackage{booktabs}
\usepackage[noadjust]{cite}
\usepackage{graphicx}

\def \NUMCLASSIFIERS {27}
\def \NUMFOLDS {10}
\def \NUMNESTED {5}
\def \NUMBAGS {10}
\def \NUMBAGGEDCLASSIFIERS {270}
\def \BRIERMETA {0.083}
\def \NEWPREDICTIONS {988}
\def \BLINDED {Pandey et al.}

\begin{document}

\title{A Comparative Analysis of Ensemble Classifiers:\\Case Studies in Genomics}

\author{
    \IEEEauthorblockN{
        Sean Whalen and Gaurav Pandey
    }
    \IEEEauthorblockA{
        Department of Genetics and Genomic Sciences \\
        Icahn Institute for Genomics and Multiscale Biology \\
        Icahn School of Medicine at Mount Sinai, New York, USA \\
        \{sean.whalen,gaurav.pandey\}@mssm.edu
    }
}

\maketitle

\begin{abstract}
The combination of multiple classifiers using ensemble methods is increasingly important for making progress in a variety of difficult prediction problems.  We present a comparative analysis of several ensemble methods through two case studies in genomics, namely the prediction of genetic interactions and protein functions, to demonstrate their efficacy on real-world datasets and draw useful conclusions about their behavior.  These methods include simple aggregation, meta-learning, cluster-based meta-learning, and ensemble selection using heterogeneous classifiers trained on resampled data to improve the diversity of their predictions.  We present a detailed analysis of these methods across 4 genomics datasets and find the best of these methods offer statistically significant improvements over the state of the art in their respective domains.  In addition, we establish a novel connection between ensemble selection and meta-learning, demonstrating how both of these disparate methods establish a balance between ensemble diversity and performance.
\end{abstract}


\section{Introduction}

Ensemble methods combining the output of individual classifiers~\cite{Rokach2009,Seni2010} have been immensely successful in producing accurate predictions for many complex classification tasks~\cite{Shotton2011,Kim1997,Yang2010,Altmann2008,Liu2012,Khan2010,Pandey2010}. The success of these methods is attributed to their ability to both consolidate accurate predictions and correct errors across many diverse base classifiers~\cite{Tumer1996}.  Diversity is key to ensemble performance: If there is complete consensus the ensemble cannot outperform the best base classifier, yet an ensemble lacking any consensus is unlikely to perform well due to weak base classifiers.  Successful ensemble methods establish a balance between the diversity and accuracy of the ensemble~\cite{Kuncheva2003,Dietterich2000}.  However, it remains largely unknown how different ensemble methods achieve this balance to extract the maximum information from the available pool of base classifiers~\cite{Kuncheva2003,Tang2006}.  A better understanding of how different ensemble methods utilize diversity to increase accuracy using complex datasets is needed, which we attempt to address with this paper.

Popular methods like bagging~\cite{Breiman1996} and boosting~\cite{Schapire2012} generate diversity by sampling from or assigning weights to training examples but generally utilize a single type of base classifier to build the ensemble. However, such {\em homogeneous} ensembles may not be the best choice for problems where the ideal base classifier is unclear.  One may instead build an ensemble from the predictions of a wide variety of {\em heterogeneous} base classifiers such as support vector machines, neural networks, and decision trees.  Two popular heterogeneous ensemble methods include a form of meta-learning called {\em stacking}~\cite{Merz1999,Wolpert1992} as well as ensemble selection~\cite{Caruana2004,Caruana2006}. Stacking constructs a higher-level predictive model over the predictions of base classifiers, while ensemble selection uses an incremental strategy to select base predictors for the ensemble while balancing diversity and performance.  Due to their ability to utilize heterogeneous base classifiers, these approaches have superior performance across several application domains~\cite{Altmann2008,Niculescu-Mizil2009}.

Computational genomics is one such domain where classification problems are especially difficult. This is due in part to incomplete knowledge of how the cellular phenomenon of interest is influenced by the variables and measurements used for prediction, as well as a lack of consensus regarding the best classifier for specific problems.  Even from a data perspective, the frequent presence of extreme class imbalance, missing values, heterogeneous data sources of different scale, overlapping feature distributions, and measurement noise further complicate classification.  These difficulties suggest that heterogeneous ensembles constructed from a large and diverse set of base classifiers, each contributing to the final predictions, are ideally suited for this domain.  Thus, in this paper we use real-world genomic datasets (detailed in Section~\ref{subsection:datasets}) to analyze and compare the performance of ensemble methods for two important problems in this area: 1) prediction of protein functions~\cite{Pandey2006}, and 2) predicting genetic interactions~\cite{Pandey2010}, both using high-throughput genomic datasets.  Constructing accurate predictive models for these problems is notoriously difficult for the above reasons, and even small improvements in predictive accuracy have the potential for large contributions to biomedical knowledge.  Indeed, such improvements uncovered the functions of mitochondrial proteins~\cite{Hess2009} and several other critical protein families.  Similarly, the computational discovery of genetic interactions between the human genes EGFR-IFIH1 and FKBP9L-MOSC2 potentially enables novel therapies for glioblastoma~\cite{Szczurek2013}, the most aggressive type of brain tumor in humans.

Working with important problems in computational genomics, we present a comparative analysis of several methods used to construct ensembles from large and diverse sets of base classifiers.  Several aspects of heterogeneous ensemble construction that have not previously been addressed are examined in detail including a novel connection between ensemble selection and meta-learning, the optimization of the diversity/accuracy tradeoff made by these disparate approaches, and the role of calibration in their performance.  This analysis sheds light on how variants of simple greedy ensemble selection achieve enhanced performance, why meta-learning often out-performs ensemble selection, and several directions for future work.  The insights obtained from the performance and behavior of ensemble methods for these complex domain-driven classification problems should have wide applicability across diverse applications of ensemble learning.

We begin by detailing our datasets, experimental methodology, and the ensemble methods studied (namely ensemble selection and stacking) in Section~\ref{section:methods}.  This is followed by a discussion of their performance in terms of standard evaluation metrics in Section~\ref{section:results}.  We next examine how the roles of diversity and accuracy are balanced in ensemble selection and establish a connection with stacking by examining the weights assigned to base classifiers by both methods (Section~\ref{subsection:diversity}).  In Section~\ref{subsection:calibration}, we discuss the impact of classifier calibration on heterogeneous ensemble performance, an important issue that has only recently received attention~\cite{Bella2012}. We conclude and indicate directions for future work in Section~\ref{section:conclusion}.

\section{Materials and Methods}
\label{section:methods}

\subsection{Problem Definitions and Datasets}
\label{subsection:datasets}

\begin{table}
\caption{
Details of genetic interaction (GI) and protein function (PF) datasets including the number of features, number of examples in the minority (positive) and majority (negative) classes, and total number of examples.
}
\centering
\begin{tabular}{lrrrr}
\toprule
Problem & Features & Positives & Negatives & Total \\
\midrule
GI & 152 & 9,994 & 125,509 & 135,503 \\
PF1 & 300 & 382 & 3,597 & 3,979 \\
PF2 & 300 & 344 & 3,635 & 3,979 \\
PF3 & 300 & 327 & 3,652 & 3,979 \\
\bottomrule
\end{tabular}
\label{table:datasets}
\end{table}

For this study we focus on two important problems in computational genomics:  The prediction of protein functions, and the prediction of genetic interactions.  Below we describe these problems and the datasets used to assess the efficacy of various ensemble methods.  A summary of these datasets is given in Table~\ref{table:datasets}.

\subsubsection{Protein Function Prediction}

A key goal in molecular biology is to infer the cellular functions of proteins. To keep pace with the rapid identification of proteins due to advances in genome sequencing technology, a large number of computational approaches have been developed to predict various types of protein functions.  These approaches use various genomic datasets to characterize the cellular functions of proteins or their corresponding genes~\cite{Pandey2006}. Protein function prediction is essentially a classification problem using features defined for each gene or its resulting protein to predict whether the protein performs a certain function ($1$) or not ($0$).  We use the gene expression compendium of Hughes et al.~\cite{Hughes2000} to predict the functions of roughly~4,000 baker's yeast (\emph{S. cerevisiae}) genes. The three most abundant functional labels from the list of Gene Ontology Biological Process terms compiled by Myers et al.~\cite{Myers2006} are used in our evaluation.  The three corresponding prediction problems are referred to as PF1, PF2, and PF3 respectively and are suitable targets for classification case studies due to their difficulty.  These datasets are publicly available from \BLINDED~\cite{Pandey2009}.

\subsubsection{Genetic Interaction Prediction}

Genetic interactions (GIs) are a category of cellular interactions that are inferred by comparing the effect of the simultaneous knockout of two genes with the effect of knocking them out individually~\cite{Hartman2001}. The knowledge of these interactions is critical for understanding cellular pathways~\cite{Horn2011}, evolution~\cite{Vandersluis2010}, and numerous other biological processes.  Despite their utility, a general paucity of GI data exists for several organisms important for biomedical research.  To address this problem, \BLINDED~\cite{Pandey2010} used ensemble classification methods to predict GIs between genes from \emph{S. cerevisiae} (baker's yeast) using functional relationships between gene pairs such as correlation between expression profiles, extent of co-evolution, and the presence or absence of physical interactions between their corresponding proteins.  We use the data from this study to assess the efficacy of heterogeneous ensemble methods for predicting GIs from a set of~152 features (see Table~\ref{table:gi_features} for an illustration) and measure the improvement of our ensemble methods over this state-of-the-art.

\begin{table}
\centering
\caption{
Feature matrix of genetic interactions where $n$ rows represent pairs of genes measured by features $F_1 \dots F_m$ having label 1 if they are known to interact, 0 if they do not, and ? if their interaction has not been established.
}
\begin{tabular}{cccccc}
\toprule
Gene Pair & F$_1$ & F$_2$ & $\cdots$ & F$_m$ & Interaction? \\
\midrule
Pair$_1$ & 0.5 & 0.1 & $\cdots$ & 0.7 & 1 \\
Pair$_2$ & 0.2 & 0.7 & $\cdots$ & 0.8 & 0 \\
$\vdots$ & $\vdots$ & $\vdots$ & $\ddots$ & $\vdots$ & $\vdots$ \\
Pair$_n$ & 0.3 & 0.9 & $\cdots$ & 0.1 & ? \\
\bottomrule
\end{tabular}
\label{table:gi_features}
\end{table}

\subsection{Experimental Setup}
\label{subsection:setup}

A total of~\NUMCLASSIFIERS\ heterogeneous classifier types are trained using the statistical language R~\cite{R} in combination with its various machine learning packages, as well as the RWeka interface~\cite{Hornik2009} to the data mining software Weka~\cite{Hall2009} (see Table~\ref{table:performance-base}).  Among these are classifiers based on boosting and bagging which are themselves a type of ensemble method, but whose performance can be further improved by inclusion in a heterogeneous ensemble. Classifiers are trained using~\NUMFOLDS-fold cross-validation where each training split is resampled with replacement~\NUMBAGS\ times then balanced using undersampling of the majority class.  The latter is a standard and essential step to prevent learning decision boundaries biased to the majority class in the presence of extreme class imbalance such as ours (see Table~\ref{table:datasets}).  In addition, a~\NUMNESTED-fold nested cross-validation is performed on each training split to create a validation set for the corresponding test split.  This validation set is used for the meta-learning and ensemble selection techniques described in Section~\ref{subsection:ensemble_methods}.  The final result is a pool of~\NUMBAGGEDCLASSIFIERS\ classifiers.

Performance is measured by combining the predictions made on each test split resulting from cross-validation into a single set and calculating the area under the Receiver Operating Characteristic curve (AUC).  The performance of the~\NUMCLASSIFIERS\ base classifiers for each dataset is given in Table~\ref{table:performance-base}, where the bagged predictions for each base classifier are averaged before calculating the AUC.  These numbers become important in later discussions since ensemble methods involve a tradeoff between the diversity of predictions and the performance of base classifiers constituting the ensemble.

\begin{table}
\centering
\caption{
Individual performance of \NUMCLASSIFIERS\ base classifiers on genetic interaction and protein function datasets evaluated using a combination of R~\cite{R}, caret~\cite{Kuhn2008}, and the RWeka interface~\cite{Hornik2009} to Weka~\cite{Hall2009}.  Details of each classifier are omitted for brevity.  R packages include a citation describing the method.  For all others, see the Weka documentation.  For boosting methods with selectable base learners, the default (normally a Decision Stump) is used.
}
\begin{tabular}{lrrrr}
& \multicolumn{4}{c}{Performance} \\
\cmidrule{2-5}

Classifier & GI & PF1 & PF2 & PF3 \\
\midrule

Functions \\
\quad glmboost~\cite{Hothorn2010} & 0.72 & 0.65 & 0.71 & 0.72 \\
\quad glmnet~\cite{Friedman2010} & 0.73 & 0.63 & 0.71 & 0.73 \\
\quad Logistic & 0.73 & 0.61 & 0.66 & 0.71 \\
\quad MultilayerPerceptron & 0.74 & 0.64 & 0.71 & 0.74 \\
\quad multinom~\cite{Venables2002} & 0.73 & 0.61 & 0.67 & 0.71 \\
\quad RBFClassifier & 0.73 & 0.62 & 0.69 & 0.74 \\
\quad RBFNetwork & 0.56 & 0.51 & 0.52 & 0.58 \\
\quad SGD & 0.73 & 0.63 & 0.70 & 0.73 \\
\quad SimpleLogistic & 0.73 & 0.65 & 0.72 & 0.73 \\
\quad SMO & 0.73 & 0.64 & 0.70 & 0.73 \\
\quad SPegasos & 0.66 & 0.52 & 0.56 & 0.56 \\
\quad VotedPerceptron & 0.65 & 0.62 & 0.70 & 0.71 \\

\addlinespace
Trees \\
\quad AdaBoostM1 & 0.71 & 0.65 & 0.67 & 0.73 \\
\quad ADTree & 0.73 & 0.64 & 0.67 & 0.75 \\
\quad gbm~\cite{Ridgeway2013} & 0.77 & \bf{0.68} & \bf{0.72} & \bf{0.78} \\
\quad J48 & 0.75 & 0.60 & 0.65 & 0.71 \\
\quad LADTree & 0.74 & 0.64 & 0.69 & 0.75 \\
\quad LMT & 0.76 & 0.62 & 0.71 & 0.75 \\
\quad LogitBoost & 0.73 & 0.65 & 0.69 & 0.75 \\
\quad MultiBoostAB & 0.70 & 0.63 & 0.66 & 0.70 \\
\quad RandomTree & 0.71 & 0.57 & 0.60 & 0.63 \\
\quad rf~\cite{Liaw2002} & \bf{0.79} & 0.67 & 0.72 & 0.76 \\

\addlinespace
Rule-Based \\
\quad JRip & 0.76 & 0.63 & 0.67 & 0.70 \\
\quad PART & 0.76 & 0.59 & 0.65 & 0.73 \\

\addlinespace
Other \\
\quad IBk & 0.70 & 0.61 & 0.66 & 0.70 \\
\quad pam~\cite{Hastie2002} & 0.71 & 0.62 & 0.66 & 0.64 \\
\quad VFI & 0.64 & 0.55 & 0.56 & 0.63 \\

\bottomrule
\end{tabular}

\label{table:performance-base}
\end{table}

\subsection{Ensemble Methods}
\label{subsection:ensemble_methods}

\subsubsection{Simple Aggregation}

The predictions of each base classifier become columns in a matrix where rows are instances and the entry at row~$i$, column~$j$ is the probability of instance~$i$ belonging to the positive class as as predicted by classifier~$j$.  We evaluate ensembles using AUC by applying the mean across rows to produce an aggregate prediction for each instance.

\subsubsection{Meta-Learning}
\label{subsection:meta-learning}

Meta-learning is a general technique for improving the performance of multiple classifiers by using the meta information they provide.  A common approach to meta-learning is stacked generalization ({\em stacking})~\cite{Wolpert1992} that trains a higher-level ({\em level~1}) classifier on the outputs of base ({\em level~0}) classifiers.

Using the standard formulation of Ting and Witten~\cite{Ting1999}, we perform meta-learning using stacking with a level~1 logistic regression classifier trained on the {\em probabilistic} outputs of multiple heterogeneous level~0 classifiers.  Though other classifiers may be used, a simple logistic regression meta-classifier helps avoid overfitting which typically results in superior performance~\cite{Ting1999}.  In addition, its coefficients have an intuitive interpretation as the weighted importance of each level~0 classifier~\cite{Altmann2008}.

The layer 1 classifier is trained on a validation set created by the nested cross-validation of a particular training split and evaluated against the corresponding test split to prevent the leaking of label information.  Overall performance is evaluated as described in Section~\ref{subsection:setup}.

In addition to stacking across all classifier outputs, we also evaluate stacking using only the {\em aggregate} output of each resampled (bagged) base classifier.  For example, the outputs of all \NUMBAGS\ SVM classifiers are averaged and used as a single level~0 input to the meta learner.  Intuitively this combines classifier outputs that have similar performance and calibration, which allows stacking to focus on weights between (instead of within) classifier types.

\subsubsection{Cluster-Based Meta-Learning}
\label{subsection:cluster_meta-learning}

A variant on traditional stacking is to first cluster classifiers with similar predictions, then learn a separate level 1 classifier for each cluster~\cite{Altmann2008}.  Alternately, classifiers within a cluster can first be combined by taking their mean (for example) and then learning a level 1 classifier on these per-cluster averaged outputs.  This is a generalization of the aggregation approach described in Section~\ref{subsection:meta-learning} but using a distance measure instead of restricting each cluster to bagged homogeneous classifiers.  We use hierarchical clustering with~\mbox{$1 - \lvert \rho \rvert $} (where~$\rho$ is Pearson's correlation) as a distance measure.  We found little difference between alternate distance measures based on Pearson and Spearman correlation and so present results using only this formulation.

For simplicity, we refer to the method of stacking within clusters and taking the mean of level~1 outputs as {\em intra-cluster stacking}.  Its complement, {\em inter-cluster stacking}, averages the outputs of classifiers within a cluster then performs stacking on the averaged level~0 outputs.  The intuition for both approaches is to group classifiers with similar (but ideally non-identical) predictions together and learn how to best resolve their disagreements via weighting.  Thus the diversity of classifier predictions within a cluster is important, and the effectiveness of this method is tied to a distance measure that can utilize both accuracy and diversity.

\subsubsection{Ensemble Selection}
\label{subsection:ensemble_selection}

Ensemble selection is the process of choosing a subset of all available classifiers that perform well together, since including every classifier may decrease performance.  Testing all possible classifier combinations quickly becomes infeasible for ensembles of any practical size and so heuristics are used to approximate the optimal subset.  The performance of the ensemble can only improve upon that of the best base classifier if the ensemble has a sufficient pool of accurate and diverse classifiers, and so successful selection methods must balance these two requirements.

We establish a baseline for this approach by performing simple greedy ensemble selection, sorting base classifiers by their individual performance and iteratively adding the best unselected classifier to the ensemble.  This approach disregards how well the classifier actually complements the performance of the ensemble.

Improving on this approach, Caruana et al.'s ensemble selection~(CES)~\cite{Caruana2004,Caruana2006} begins with an empty ensemble and iteratively adds new predictors that maximize its performance according to a chosen metric (here, AUC).  At each iteration, a number of candidate classifiers are randomly selected and the performance of the current ensemble including the candidate is evaluated.  The candidate resulting in the best ensemble performance is selected and the process repeats until a maximum ensemble size is reached.  The evaluation of candidates according to their performance with the ensemble, instead of in isolation, improves the performance of CES over simple greedy selection.

Additional improvements over simple greedy selection include ~1) initializing the ensemble with the top $n$ base classifiers, and~2) allowing classifiers to be added multiple times.  The latter is particularly important as without replacement, the best classifiers are added early and ensemble performance then decreases as poor predictors are forced into the ensemble.  Replacement gives more weight to the best performing predictors while still allowing for diversity.  We use an initial ensemble size of~\mbox{$n = 2$} to reduce the effect of multiple bagged versions of a single high performance classifier dominating the selection process, and (for completeness) evaluate all candidate classifiers instead of sampling.

Ensemble predictions are combined using a cumulative moving average to speed the evaluation of ensemble performance for each candidate predictor.  Selection is performed on the validation set produced by nested cross-validation and the resulting ensemble evaluated as described in Section~\ref{subsection:setup}.

\subsection{Diversity Measures}

 The diversity of predictions made by members of an ensemble determines the ensemble's ability to outperform the best individual, a long-accepted property which we explore in the following sections.  We measure diversity using Yule's $Q$-statistic~\cite{Yule1900} by first creating predicted labels from thresholded classifier probabilities, yielding a~1 for values greater than~0.5 and~0 otherwise.  Given the predicted labels produced by each pair of classifiers~$D_i$ and~$D_k$, we generate a contingency table counting how often each classifier produces the correct label in relation to the other:

\begin{center}
\begin{tabular}{lcc}
& $D_k$ correct (1) & $D_k$ incorrect (0) \\
\midrule
$D_i$ correct (1) & $N^{11}$ & $N^{10}$ \\
$D_i$ wrong (0) & $N^{01}$ & $N^{00}$ \\
\end{tabular}
\end{center}

\noindent The pairwise~$Q$ statistic is then defined as:
\begin{align}
Q_{i,k} = \frac{N^{11} N^{00} - N^{01} N^{10}}{N^{11} N^{00} + N^{01} N^{10}} ~.
\end{align}

\noindent This produces values tending towards~$1$ when~$D_i$ and~$D_k$ correctly classify the same instances,~$0$ when they do not, and~$-1$ when they are negatively correlated.  We evaluated additional diversity measures such as Cohen's~$\kappa$-statistic~\cite{Cohen1960} but found little practical difference between the measures (in agreement with Kuncheva et al.~\cite{Kuncheva2003}) and focus on~$Q$ for its simplicity.  Multicore performance and diversity measures are implemented in C++ using the Rcpp package~\cite{Eddelbuettel2011}.  This proves essential for their practical use with large ensembles and nested cross validation.

We adjust raw~$Q$ values using the transformation~\mbox{$1 - \lvert Q \rvert $} so that~0 represents no diversity and~1 represents maximum diversity for graphical clarity.

\section{Ensemble Performance}
\label{section:results}

Performance of the methods described in Section~\ref{subsection:ensemble_methods} is summarized in Table~\ref{table:performance-ensemble}.  Overall, aggregated stacking is the best performer and edges out CES for all our datasets.  The use of clustering in combination with stacking also performs well for certain cluster sizes~$k$.  Intra-cluster stacking performs best with cluster sizes~2,~14,~20, and~15 for GI, PF1, PF2, and PF3, respectively.  Inter-cluster stacking is optimal for sizes~24,~33,~33, and~36 on the same datasets.  Due to the size of the GI dataset, only~10\% of the validation set is used for non-aggregate stacking and cluster stacking methods.  Performance levels off beyond 10\% and so this approach does not significantly penalize these methods.  This step was not necessary for the other methods and datasets.

\begin{table}
\caption{
AUC of ensemble learning methods for protein function and genetic interaction datasets.  Methods include mean aggregation, greedy ensemble selection, selection with replacement (CES), stacking with logistic regression, aggregated stacking (averaging resampled homogeneous base classifiers before stacking), stacking within clusters then averaging (intra), and averaging within clusters then stacking (intra).  The best performing base classifier (random forest for the GI dataset and GBM for PFs) is given for reference.  Starred values are generated from a subsample of the validation set due to its size; see text for detail.
}
\centering
\begin{tabular}{lrrrr}
& \multicolumn{4}{c}{Performance} \\
\cmidrule{2-5}
Method & GI & PF1 & PF2 & PF3 \\
\midrule
Best Base Classifier & 0.79\phantom{$^*$} & 0.68 & 0.72 & 0.78 \\
Mean Aggregation & 0.763\phantom{$^*$} & 0.669 & 0.732 & 0.773 \\
Greedy Selection & 0.792\phantom{$^*$} & 0.684 & 0.734 & 0.779 \\
CES & 0.802\phantom{$^*$} & 0.686 & 0.741 & 0.785 \\
Stacking (Aggregated) & \bf{0.812}\phantom{$^*$} & \bf{0.687} & \bf{0.742} & \bf{0.788} \\
Stacking (All) & 0.809$^*$ & 0.684 & 0.726 & 0.773 \\
Intra-Cluster Stacking & 0.799$^*$ & 0.684 & 0.725 & 0.775 \\
Inter-Cluster Stacking & 0.786$^*$ & 0.683 & 0.735 & 0.783 \\
\bottomrule
\end{tabular}

\label{table:performance-ensemble}
\end{table}

\begin{table}
\caption{Pairwise performance comparison of multiple non-ensemble and ensemble methods across datasets.  Only pairs with statistically significant differences, determined by Friedman/Nemenyi tests at~\mbox{$\alpha = 0.05$}, are shown.}
\centering
\begin{tabular}{llr}
Method A & Method B & p-value \\
\midrule
Best Base Classifier & CES & 0.001902 \\
Best Base Classifier & Stacking (Aggregated) & 0.000136 \\
CES & Inter-Cluster Stacking & 0.037740 \\
CES & Intra-Cluster Stacking & 0.014612 \\
CES & Mean Aggregation & 0.000300 \\
CES & Stacking (All) & 0.029952 \\
Greedy Selection & Mean Aggregation & 0.029952 \\
Greedy Selection & Stacking (Aggregated) & 0.005364 \\
Inter-Cluster Stacking & Mean Aggregation & 0.047336 \\
Inter-Cluster Stacking & Stacking (Aggregated) & 0.003206 \\
Intra-Cluster Stacking & Stacking (Aggregated) & 0.001124 \\
Mean Aggregation & Stacking (Aggregated) & 0.000022 \\
Stacking (Aggregated) & Stacking (All) & 0.002472 \\
\end{tabular}

\label{table:friedman-pairwise}
\end{table}

\begin{table}
\caption{Grouped performance comparison of multiple non-ensemble and ensemble methods across datasets.  Methods sharing a group letter have statistically similar performance, determined by Friedman/Nemenyi tests at~\mbox{$\alpha = 0.05$}.  Aggregated stacking and CES demonstrate the best performance, while Greedy selection is similar to CES (but not stacking).}
\centering
\begin{tabular}{llr}
Group & Method & Rank Sum \\
\midrule
a    & Stacking (Aggregated)        & 32 \\
ab   & CES                          & 27 \\
bc   & Greedy Selection             & 18 \\
c    & Inter-Cluster Stacking       & 17 \\
cd   & Stacking (All)               & 16.5 \\
cd   & Intra-Cluster Stacking       & 15 \\
cd   & Best Base Classifier         & 11 \\
d    & Mean Aggregation             & 7.5 \\
\end{tabular}

\label{table:friedman-grouped}
\end{table}

Ensemble selection also performs well, though we anticipate issues of calibration (detailed in Section~\ref{subsection:calibration}) could have a negative impact since the mean is used to aggregate ensemble predictions.  Greedy selection achieves best performance for ensemble sizes of~10,~14,~45, and~38 for GI, PF1, PF2, and PF3, respectively.  CES is optimal for sizes~70,~43,~34, and~56 for the same datasets.  Though the best performing ensembles for both selection methods are close in performance, simple greedy selection is much worse {\em for non-optimal ensemble sizes} than CES and its performance typically degrades after the best few base classifiers are selected (see Section~\ref{subsection:diversity}).  Thus, on average CES is the superior selection method.

In agreement with Altman et al.~\cite{Altmann2008}, we find the mean is the highest performing simple aggregation method for combining ensemble predictions.  However, because we are using heterogeneous classifiers that may have uncalibrated outputs, the mean combines predictions made with different scales or notions of probability.  This explains its poor performance compared to the best base classifier in a heterogeneous ensemble and emphasizes the need for ensemble selection or weighting via stacking to take full advantage of the ensemble.  We discuss the issue of calibration in Section~\ref{subsection:calibration}.

Thus, we observe consistent performance trends across these methods.  However, to draw meaningful conclusions it is critical to determine if the performance differences are statistically significant.  For this we employ the standard methodology given by Dem\v{s}ar~\cite{Demsar2006} to test for statistically significant performance differences between multiple methods across multiple datasets.  The Friedman test~\cite{Friedman1937} first determines if there are statistically significant differences between any pair of methods over all datasets, followed by a post-hoc Nemenyi test~\cite{Nemenyi1963} to calculate a p-value for each pair of methods.  This is the non-parametric equivalent of ANOVA combined with a Tukey HSD post-hoc test where the assumption of normally distributed values is removed by using rank transformations.  As many of the assumptions of parametric tests are violated by machine learning algorithms, the Friedman/Nemenyi test is preferred despite reduced statistical power~\cite{Demsar2006}.

Using the Freidman/Nemeyi approach with a cutoff of~\mbox{$\alpha = 0.05$}, the pairwise comparison between our ensemble and non-ensemble methods is shown in Table~\ref{table:friedman-pairwise}.  For brevity, only methods with statistically significant performance differences are shown.  The ranked performance of each method across all datasets is shown in Table~\ref{table:friedman-grouped}.  Methods sharing a label in the group column have statistically indistinguishable performance based on their summed rankings.  This table shows that aggregated stacking and CES have the best performance, while CES and pure greedy selection have similar performance.  However, aggregated stacking and greedy selection do not share a group as their summed ranks are too distant and thus have a significant performance difference.  The remaining approaches including non-aggregated stacking are statistically similar to mean aggregation and motivates our inclusion of cluster-based stacking, whose performance may improve given a more suitable distance metric.  These rankings statistically reinforce the general trends presented earlier in Table~\ref{table:performance-ensemble}.

\begin{figure}
\centering
\includegraphics[width=\columnwidth]{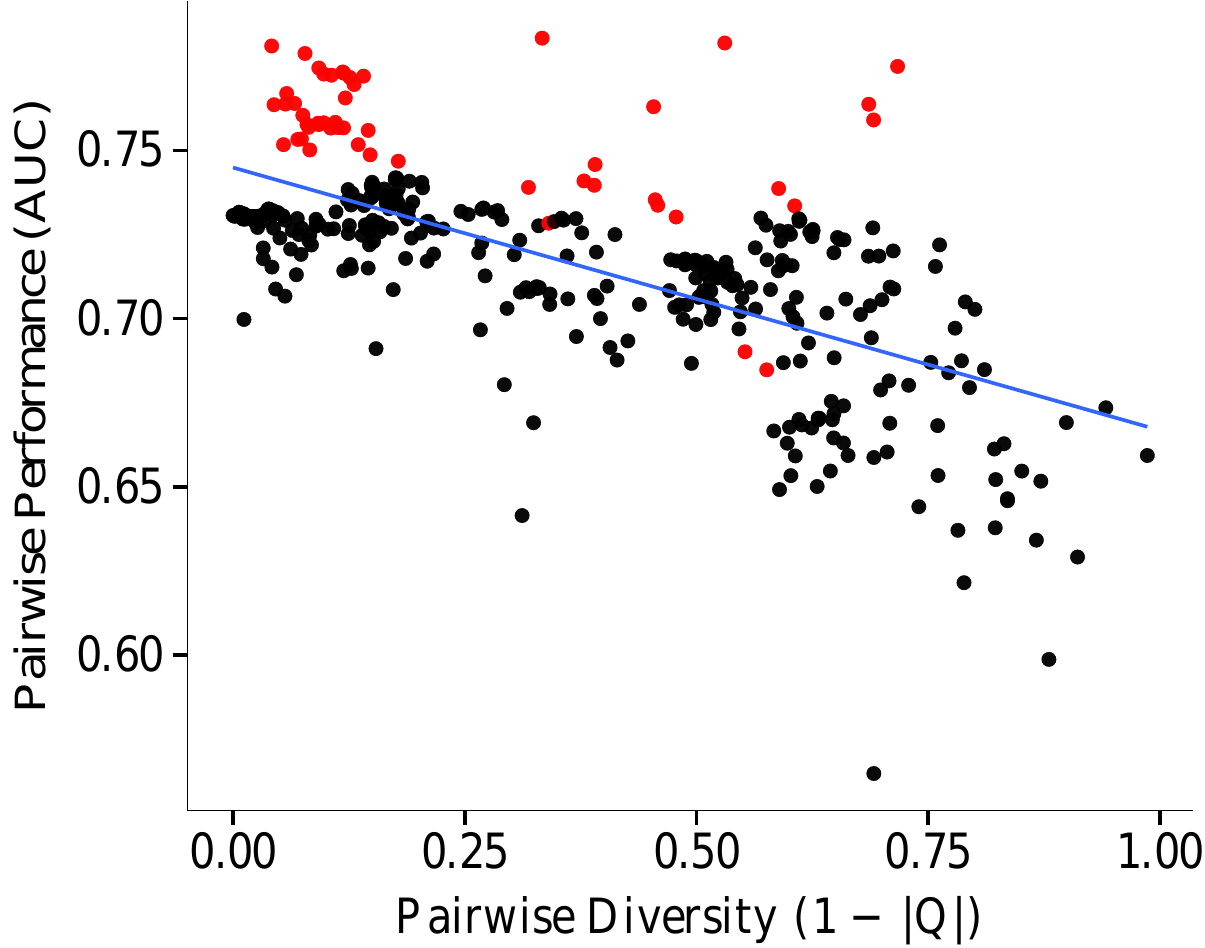}
\phantom{bla}
\includegraphics[width=\columnwidth]{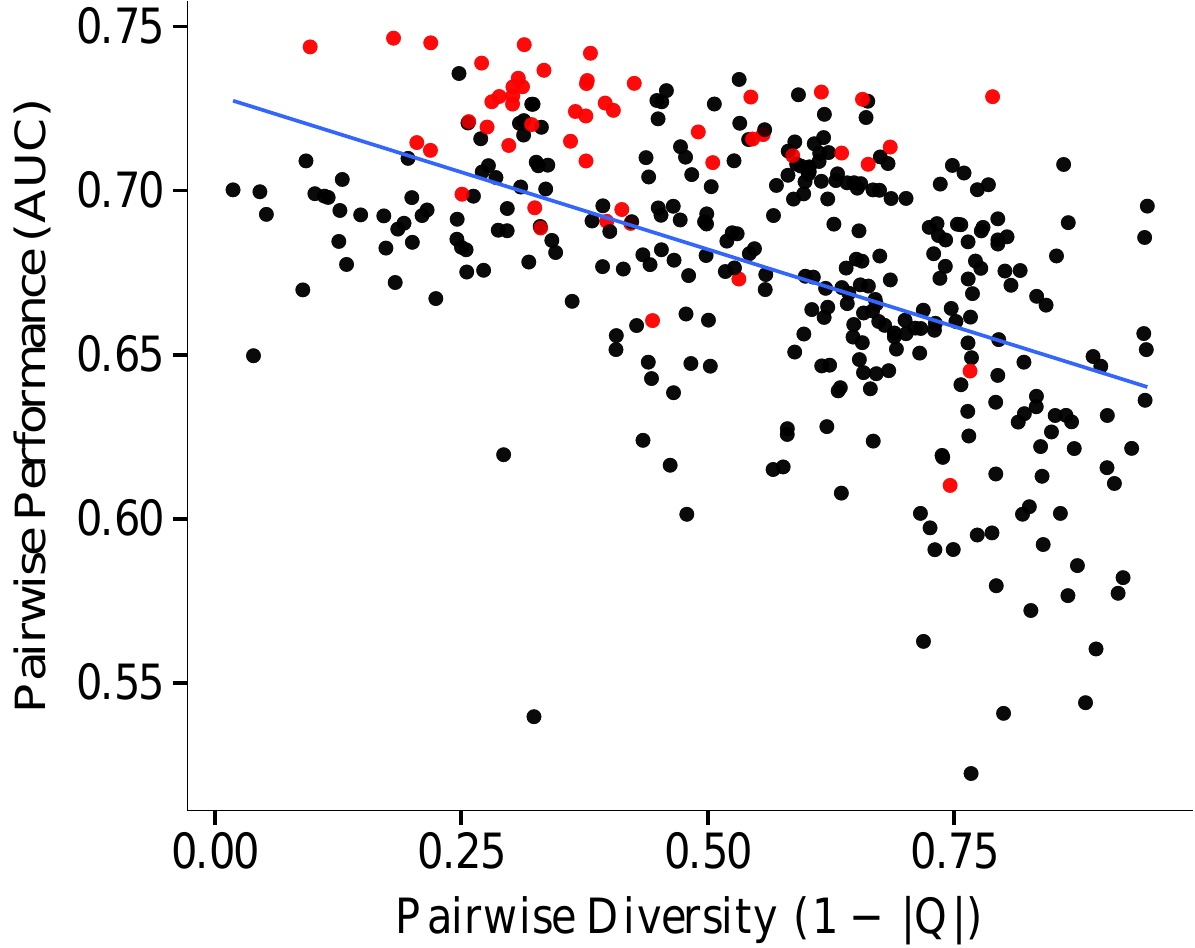}
\caption{
Performance as a function of diversity for all pairwise combinations of \NUMCLASSIFIERS\ base classifiers on the GI (top figure) and PF3 (bottom figure) datasets.  More diverse combinations typically result in less performance except for high-performance classifiers such as random forests and generalized boosted regression models, whose points are shown in red if they are part of a pair.  Raw $Q$ values are adjusted so that larger values imply more diversity.
}
\label{figure:ensemble-diversity}
\end{figure}

We note that nested cross-validation, relative to a single validation set, improves the performance of both stacking and CES by increasing the amount of meta data available as well as the bagging that occurs as a result.  Both effects reduce overfitting but performance is still typically better with smaller ensembles.  More nested folds increase the quality of the meta data and thus affects the performance of these methods as well, though computation time increases substantially and motivates our selection of~\mbox{$k = \NUMNESTED$} nested folds.

Finally, we emphasize that each method we evaluate out-performs the previous state of the art AUC of~0.741 for GI prediction~\cite{Pandey2010}.  In particular, stacked aggregation results in the prediction of~\NEWPREDICTIONS\ additional genetic interactions at a~10\% false discovery rate.  In addition, these heterogeneous ensemble methods out-perform random forests and gradient boosted regression models which are themselves homogeneous ensembles.  This demonstrates the value of heterogeneous ensembles for improving predictive performance.

\section{Ensemble Characteristics}
\label{section:analysis}

\subsection{The Role of Diversity}
\label{subsection:diversity}

\begin{figure}
\centering
\includegraphics[width=\columnwidth]{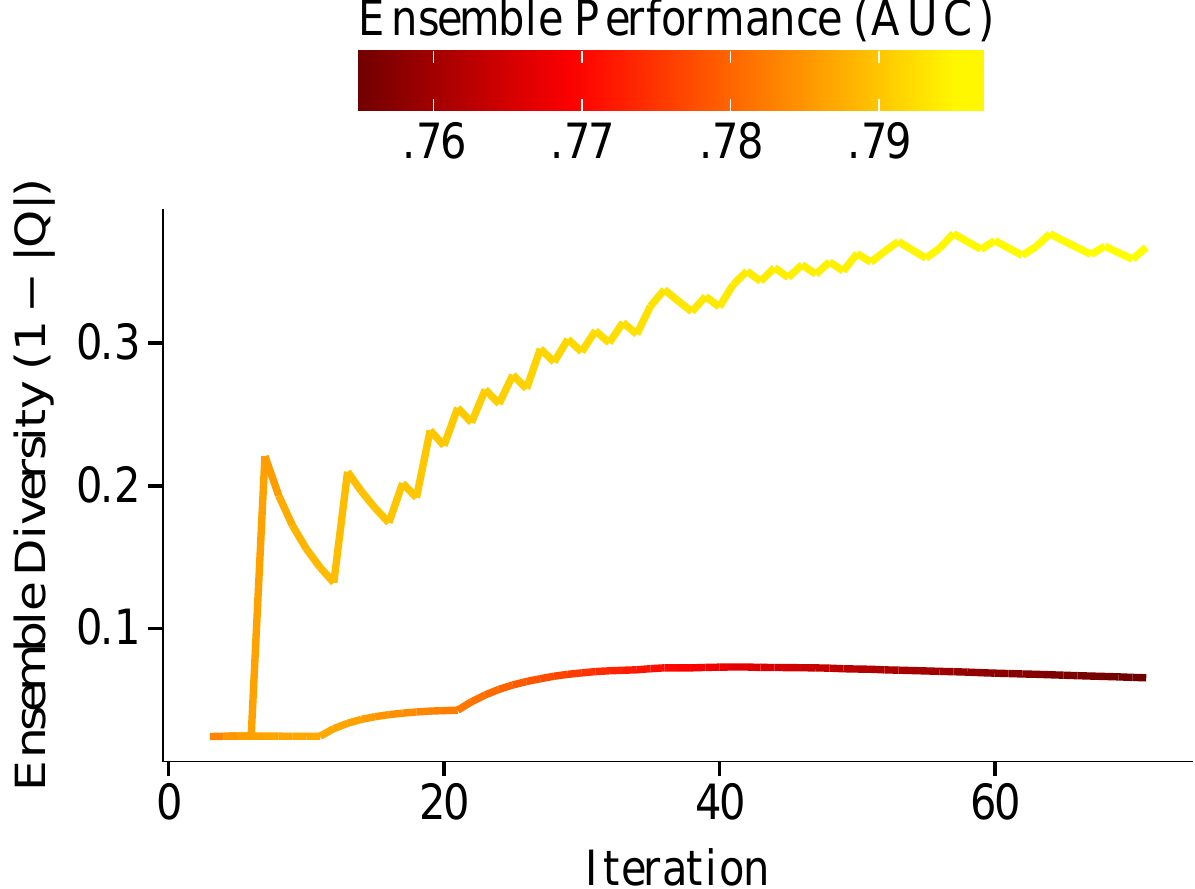}
\phantom{bla}
\includegraphics[width=\columnwidth]{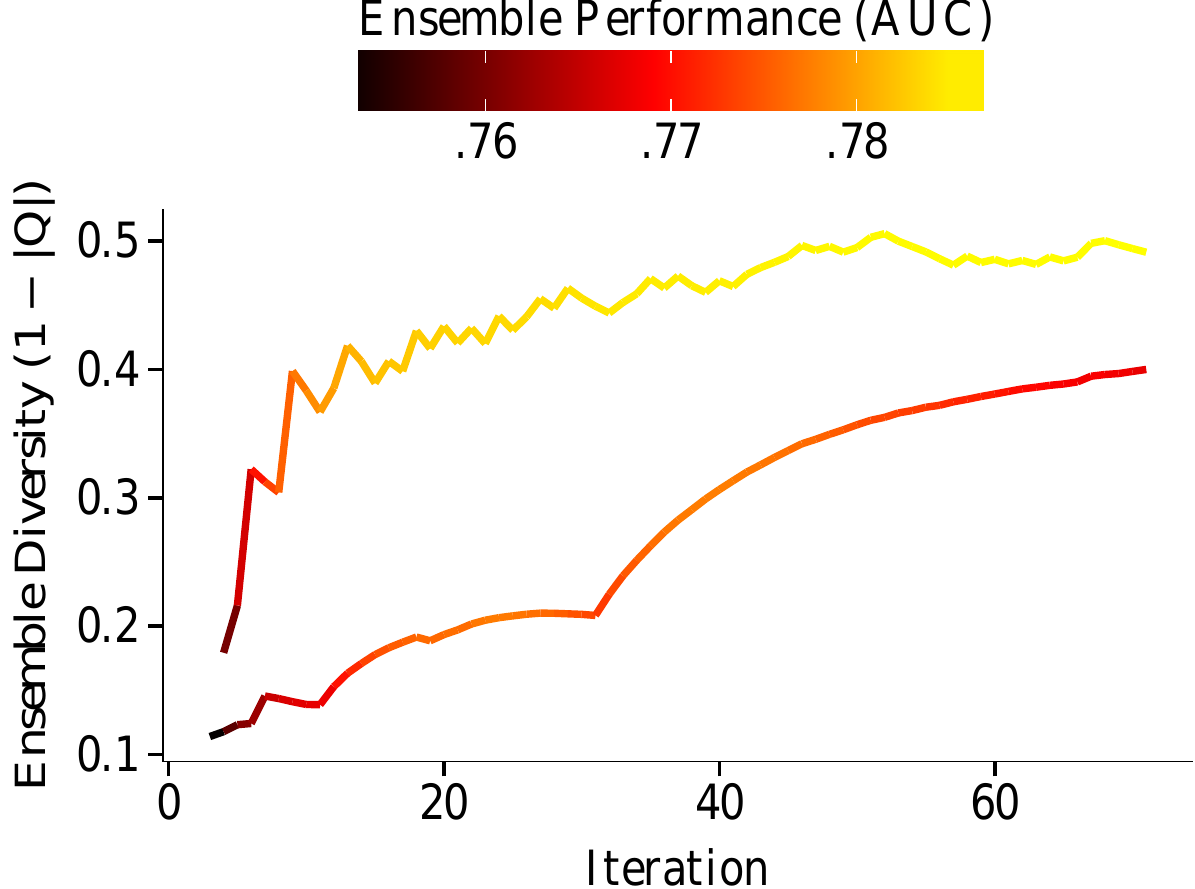}
\caption{
Ensemble diversity and performance as a function of iteration number for both greedy selection (bottom curve) and CES (top curve) on the GI and PF3 datasets.  For GI (top figure), greedy selection fails to improve diversity and performance decreases with ensemble size, while CES successfully balances diversity and performance (shown by a shift in color from red to yellow as height increases) and reaches an equilibrium over time.  For PF3 (bottom figure), greedy selection manages to improve diversity around iteration 30 but accuracy decreases, demonstrating that diverse predictions alone are not enough for accurate ensembles.
}
\label{figure:selection}
\end{figure}

The relationship between ensemble diversity and performance has immediate impact on practitioners of ensemble methods, yet has not formally been proven despite extensive study~\cite{Kuncheva2003,Tang2006}.  For brevity we analyze this tradeoff using GI and PF3 as representative datasets, though the trends observed generalize to PF1 and PF2.

Figure~\ref{figure:ensemble-diversity} presents a high-level view of the relationship between performance and diversity, plotting the diversity of pairwise classifiers against their performance as an ensemble by taking the mean of their predictions.  This figure shows the complicated relationship between diversity and performance that holds for each of our datasets: Two highly diverse classifiers are more likely to perform poorly due to lower prediction consensus.  There are exceptions, and these tend to include well-performing base classifiers such as random forests and gradient boosted regression models (shown in red in Figure~\ref{figure:ensemble-diversity}) which achieve high AUC on their own and stand to gain from a diverse partner.  Diversity works in tension with performance, and while improving performance depends on diversity, the wrong kind of diversity limits performance of the ensemble~\cite{Brown2010}.

Figure~\ref{figure:selection} demonstrates this tradeoff by plotting ensemble diversity and performance as a function of the iteration number of the simple greedy selection and CES methods detailed in Section~\ref{section:methods} for the GI (top figure) and PF3 (bottom figure) datasets.  These figures reveal how CES (top curve) successfully exploits the tradeoff between diversity and performance while a purely greedy approach (bottom curve) actually decreases in performance over iterations after the best individual base classifiers are added.  This is shown via coloring, where CES shifts from red to yellow (better performance) as its diversity increases while greedy selection grows darker red (worse performance) as its diversity only slightly increases.  Note that while greedy selection increases ensemble diversity around iteration 30 for PF3, overall performance continues to decrease.  This demonstrates that diversity must be balanced with accuracy to create well-performing ensembles.

To illustrate using the PF3 panel of Figure~\ref{figure:selection}, the first classifiers chosen by CES (in order) are rf.1, rf.7, gbm.2, RBFClassifier.0, MultilayerPerceptron.9, and gbm.3 where numbers indicate bagged versions of a base classifier.  RBFClassifier.0 is a low performance, high diversity classifier while the others are the opposite (see Table~\ref{table:performance-base} for a summary of base classifiers).  This ensemble shows how CES tends to repeatedly select base classifiers that improve performance, then selects a more diverse and typically worse performing classifier.  Here the former are different bagged versions of a random forest while the latter is RBFClassifier.0.  This manifests in the left part of the upper curve where diversity is low and then jumps to its first peak.  After this, a random forest is added again to balance performance and diversity drops until the next peak.  This process is repeated while the algorithm approaches a weighted equilibrium of high performing, low diversity and low performing, high diversity classifiers.

This agrees with recent observations that diversity enforces a kind of regularization for ensembles~\cite{Tang2006,Li2012}: Performance stops increasing when there is no more diversity to extract from the pool of possible classifiers.  We see this in the PF3 panel of Figure~\ref{figure:selection} as performance reaches its peak, where small oscillations in diversity represent re-balancing the weights to maintain performance past the optimal ensemble size.

\begin{table}
\centering
\caption{
The most-weighted classifiers produced by stacking with logistic regression (Weight$^m$) and CES (Weight$^c$) for the PF3 dataset, along with their average pairwise diversity and performance.
}
\begin{tabular}{lrrrr}
\toprule
Classifier & Weight$^m$ & Weight$^c$ & Div. & AUC \\
\midrule
  rf & 0.25 & 0.21 & 0.39  & 0.71 \\
  gbm & 0.20 & 0.27 & 0.42 & 0.72 \\
  RBFClassifier & - & 0.05 & 0.45 & 0.71 \\
  \small{MultilayerPerceptron} & 0.09 & - & 0.46 & 0.70 \\
  SGD & 0.09 & 0.04 & 0.47 & 0.69 \\
  VFI & 0.11 & 0.11 & 0.71 & 0.66 \\
  IBk & 0.09 & 0.13 & 0.72 & 0.68 \\

\bottomrule
\end{tabular}
\label{table:selection-weights}
\end{table}

Since ensemble selection and stacking are top performers and can both be interpreted as learning to weight different base classifiers, we next compare the most heavily weighted classifiers selected by CES (Weight$^c$) with the coefficients of a level~1 logistic regression meta-learner (Weight$^m$).  We compute Weight$^c$ as the normalized counts of classifiers included in the ensemble, resulting in greater weight for classifiers selected multiple times.  These weights for PF3 are shown in Table~\ref{table:selection-weights}.

Nearly the same classifiers receive the most weight under both approaches (though logistic regression coefficients were not restricted to positive values so we cannot directly compare weights between methods).  However, the general trend of the relative weights is clear and explains the oscillations seen in Figure~\ref{figure:selection}: High performance, low diversity classifiers are repeatedly paired with higher diversity, lower performing classifiers.  A more complete picture of selection emerges by examining the full list of candidate base classifiers (Table~\ref{table:performance-diversity}) with the most weighted ensemble classifiers shown in bold.  The highest performing, lowest diversity GBM and RF classifiers appear at the top of the list while VFI and IBk are near the bottom.  Though there are more diverse classifiers than VFI and IBk, they were not selected due to their lower performance.

\begin{table}
\centering
\caption{
Candidate classifiers sorted by mean pairwise diversity and performance.  The most heavily weighted classifiers for both CES and stacking are shown in bold.  This trend, which holds across datasets, shows the pairing of high-performance low-diversity classifiers with their complements, demonstrating how seemingly disparate approaches create a balance of diversity and performance.
}
\begin{tabular}{lrr}
\toprule
Classifier & Diversity & AUC \\
  \midrule
  \bf{rf} & \bf{0.386} & \bf{0.712} \\
  \bf{gbm} & \bf{0.419} & \bf{0.720} \\
  glmnet & 0.450 & 0.694 \\
  glmboost & 0.452 & 0.680 \\
  \bf{RBFClassifier} & \bf{0.453} & \bf{0.713} \\
  SimpleLogistic & 0.459 & 0.689 \\
  \bf{MultilayerPerceptron} & \bf{0.459} & \bf{0.706} \\
  SMO & 0.462 & 0.695 \\
  \bf{SGD} & \bf{0.470} & \bf{0.693} \\
  LMT & 0.472 & 0.692 \\
  pam & 0.525 & 0.673 \\
  LogitBoost & 0.528 & 0.691 \\
  ADTree & 0.539 & 0.682 \\
  VotedPerceptron & 0.540 & 0.694 \\
  multinom & 0.553 & 0.677 \\
  LADTree & 0.568 & 0.678 \\
  Logistic & 0.574 & 0.669 \\
  AdaBoostM1 & 0.584 & 0.684 \\
  PART & 0.615 & 0.652 \\
  MultiBoostAB & 0.615 & 0.679 \\
  J48 & 0.632 & 0.645 \\
  JRip & 0.687 & 0.627 \\
  \bf{VFI} & \bf{0.713} & \bf{0.662} \\
  \bf{IBk} & \bf{0.720} & \bf{0.682} \\
  RBFNetwork & 0.778 & 0.653 \\
  RandomTree & 0.863 & 0.602 \\
  SPegasos & 0.980 & 0.634 \\
   \bottomrule

\end{tabular}
\label{table:performance-diversity}
\end{table}

This example illustrates how diversity and performance are balanced during selection, and also gives new insight into the nature of stacking due to the convergent weights of these seemingly different approaches.  A metric incorporating both measures should increase the performance of hybrid methods such as cluster-based stacking, which we plan to investigate in future work.

\begin{figure*}
\centering
\includegraphics[width=\linewidth]{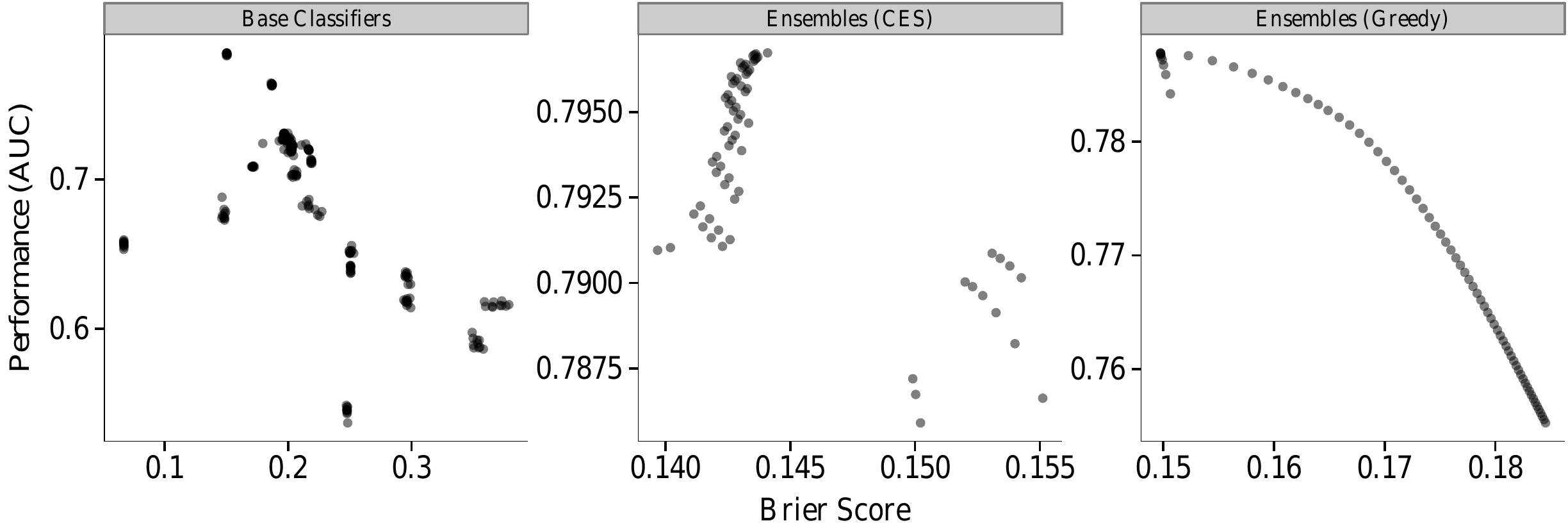}
\caption{
Performance as a function of calibration as measured by the Brier score~\cite{Brier1950} for base classifiers (left), ensembles for each iteration of CES (middle), and ensembles for each strictly greedy iteration (right).  Iterations for greedy selection move from the upper left to the lower right, while CES starts in the lower right and moves to the upper left.  This shows better-calibrated (lower Brier score) classifiers and ensembles have higher average performance and illustrates the iterative performance differences between the methods.  Stacking with logistic regression produces outputs with approximately half the best Brier score of CES, explaining the difference in final classifier weights leading to the superior performance of stacking.
}
\label{figure:ensemble-brier_score}
\end{figure*}

\subsection{The Role of Calibration}
\label{subsection:calibration}

A key factor in the performance difference between stacking and CES is illustrated by stacking's selection of MultilayerPerceptron instead of RBFClassifier for PF3.  This difference in the relative weighting of classifiers, or the exchange of one classifier for another in the final ensemble, persists across our datasets.  We suggest this is due to the ability of the layer 1 classifier to learn a function on the probabilistic outputs of base classifiers and compensate for potential differences in {\em calibration}, resulting in the superior performance of stacking.

A binary classifier is said to be {\em well-calibrated} if it is correct~$p$ percent of the time for predictions of confidence~$p$~\cite{Bella2012}.  However, accuracy and calibration are related but not the same: A binary classifier that flips a fair coin for a balanced dataset will be calibrated but not accurate.  Relatedly, many well-performing classifiers do not produce calibrated probabilities.  Measures such as AUC are not sensitive to calibration for base classifiers, and the effects of calibration on heterogeneous ensemble learning have only recently been studied~\cite{Bella2012}.  This section further investigates this relationship.

To illustrate a practical example of calibration, consider a support vector machine.  An uncalibrated SVM outputs the distance of an instance from a hyperplane to generate a probability.  This is not a true posterior probability of an instance belonging to a class, but is commonly converted to such using Platt's method~\cite{Platt1999}.  In fact, this is analogous to fitting a layer 1 logistic regression to the uncalibrated SVM outputs with a slight modification to avoid overfitting.  This approach is not restricted to SVMs and additional methods such as isotonic regression are commonly used for both binary and multi-class problems~\cite{Zadrozny2002}.

Regardless of the base classifier, a lack of calibration may effect the performance of ensemble selection methods such as CES since the predictions of many heterogeneous classifiers are combined using simple aggregation methods such as the mean.  Several methods exist for evaluating the calibration of probabilistic classifiers.  One such method, the {\em Brier score}, assesses how close (on average) a classifier's probabilistic output is to the correct binary label~\cite{Brier1950}:
\begin{align}
BS = \frac{1}{N} \sum_{i=1}^{N} (f_i - o_i)^2
\end{align}

\noindent over all instances~$i$.  This is simply the mean squared error evaluated in the context of probabilistic binary classification.  Lower scores indicate better calibration.

Figure~\ref{figure:ensemble-brier_score} plots the Brier scores for each base classifier against its performance for the GI dataset as well as the ensemble Brier scores for each iteration of CES and greedy selection.  This shows that classifiers and ensembles with calibrated outputs generally perform better.  Note in particular the calibration and performance of simple greedy selection, with initial iterations in the upper left of the panel showing high performing well-calibrated base classifiers chosen for the ensemble, but moving to the lower right as sub-optimal classifiers are forced into the ensemble.  In contrast, CES starts with points in the lower right and moves to the upper left as both ensemble calibration and performance improve each iteration.  The upper left of the CES plot suggests the benefit of additional classifiers outweighs a loss in calibration during its final iterations.

Stacking produces a layer 1 classifier with approximately half the Brier score~(\BRIERMETA) of CES or the best base classifiers.  Since this approach learns a function over probabilities it is able to adjust to the different scales used by potentially ill-calibrated classifiers in a heterogeneous ensemble.  This explains the difference in the final weights assigned by stacking and CES to the base classifiers in Table~\ref{table:selection-weights}: Though the relative weights are mostly the same, logistic regression is able to correct for the lack of calibration across classifiers and better incorporate the predictions of MultilayerPerceptron whereas CES cannot.  In this case, a calibrated MultilayerPerceptron serves to improve performance of the ensemble and thus stacking outperforms CES.

In summary, this section demonstrates the tradeoff between performance and diversity made by CES and examines its connection with stacking.  There is significant overlap in the relative weights of the most important base classifiers selected by both methods.  From this set of classifiers, stacking often assigns more weight to a particular classifier as compared to CES and this result holds across our datasets.  We attribute the superior performance of stacking to this difference, originating from its ability to accommodate differences in classifier calibration that are likely to occur in large heterogeneous ensembles. This claim is substantiated by its significantly lower Brier score compared to CES as well as the correlation between ensemble calibration and performance.  This suggests the potential for improving ensemble methods by accommodating differences in calibration.

\section{Conclusions and Future Work}
\label{section:conclusion}

The aim of ensemble techniques is to combine diverse classifiers in an intelligent way such that the predictive accuracy of the ensemble is greater than that of the best base classifier.  Since enumerating the space of all classifier combinations quickly becomes infeasible for even relatively small ensemble sizes, other methods for finding well performing ensembles have been widely studied and applied in the last decade.

In this paper we apply a variety of ensemble approaches to two difficult problems in computational genomics: The prediction of genetic interactions and the prediction of protein functions. These problems are notoriously difficult for their extreme class imbalance, prevalence of missing values, integration of heterogeneous data sources of different scale, and overlap between feature distributions of the majority and minority classes.  These issues are amplified by the inherent complexity of the underlying biological mechanisms and incomplete domain knowledge.

We find that stacking and ensemble selection approaches offer statistically significant improvements over the previous state-of-the-art for GI prediction~\cite{Pandey2010} and moderate improvements over tuned random forest classifiers which are particularly effective in this domain~\cite{Yang2010}.  Here, even small improvements in accuracy can contribute directly to biomedical knowledge after wet-lab verification: These include~\NEWPREDICTIONS\ additional genetic interactions predicted by aggregated stacking at a~10\% false discovery rate.  We also uncover a novel connection between stacking and Caruana et al.'s ensemble selection method (CES)~\cite{Caruana2004,Caruana2006}, demonstrating how these two disparate methods converge to nearly the same final base classifier weights by balancing diversity and performance in different ways.  We explain how variations in these weights are related to the calibration of base classifiers in the ensemble, and finally describe how stacking improves accuracy by accounting for differences in calibration.  This connection also shows how the utilization of diversity is an emergent, not explicit, property of how CES maximizes ensemble performance and suggests directions for future work including formalizing the effects of calibration on heterogeneous ensemble performance, modifications to CES which explicitly incorporate diversity~\cite{Li2012}, and an optimization-based formulation of the diversity/performance tradeoff for improving cluster-based stacking methods.

\section{Acknowledgements}

We thank the Genomics Institute at Mount Sinai for their generous financial and technical support.

\bibliographystyle{IEEEtran}
\bibliography{icdm2013}

\begin{thebibliography}{10}
\providecommand{\url}[1]{#1}
\csname url@samestyle\endcsname
\providecommand{\newblock}{\relax}
\providecommand{\bibinfo}[2]{#2}
\providecommand{\BIBentrySTDinterwordspacing}{\spaceskip=0pt\relax}
\providecommand{\BIBentryALTinterwordstretchfactor}{4}
\providecommand{\BIBentryALTinterwordspacing}{\spaceskip=\fontdimen2\font plus
\BIBentryALTinterwordstretchfactor\fontdimen3\font minus
  \fontdimen4\font\relax}
\providecommand{\BIBforeignlanguage}[2]{{%
\expandafter\ifx\csname l@#1\endcsname\relax
\typeout{** WARNING: IEEEtran.bst: No hyphenation pattern has been}%
\typeout{** loaded for the language `#1'. Using the pattern for}%
\typeout{** the default language instead.}%
\else
\language=\csname l@#1\endcsname
\fi
#2}}
\providecommand{\BIBdecl}{\relax}
\BIBdecl

\bibitem{Rokach2009}
L.~Rokach, ``{Ensemble-Based Classifiers},'' \emph{Artificial Intelligence
  Review}, vol.~33, no. 1-2, pp. 1--39, 2009.

\bibitem{Seni2010}
G.~Seni and J.~Elder, \emph{{Ensemble Methods in Data Mining: Improving
  Accuracy Through Combining Predictions}}.\hskip 1em plus 0.5em minus
  0.4em\relax Morgan \& Claypool, 2010.

\bibitem{Shotton2011}
J.~Shotton, A.~Fitzgibbon, M.~Cook, T.~Sharp, M.~Finocchio, R.~Moore,
  A.~Kipman, and A.~Blake, ``\BIBforeignlanguage{English}{{Real-Time Human Pose
  Recognition in Parts from Single Depth Images}},'' in
  \emph{\BIBforeignlanguage{English}{Proceedings of the 2011 IEEE Conference on
  Computer Vision and Pattern Recognition}}, 2011, pp. 1297--1304.

\bibitem{Kim1997}
D.~Kim, ``\BIBforeignlanguage{English}{{Forecasting Time Series with Genetic
  Fuzzy Predictor Ensemble}},'' \emph{\BIBforeignlanguage{English}{IEEE
  Transactions on Fuzzy Systems}}, vol.~5, no.~4, pp. 523--535, 1997.

\bibitem{Yang2010}
P.~Yang, Y.~H. Yang, B.~B. Zhou, and A.~Y. Zomaya, ``{A Review of Ensemble
  Methods in Bioinformatics},'' \emph{Current Bioinformatics}, vol.~5, no.~4,
  pp. 296--308, 2010.

\bibitem{Altmann2008}
A.~Altmann, M.~Rosen-Zvi, M.~Prosperi, E.~Aharoni, H.~Neuvirth,
  E.~Sch\"{u}lter, J.~B\"{u}ch, D.~Struck, Y.~Peres, F.~Incardona,
  A.~S\"{o}nnerborg, R.~Kaiser, M.~Zazzi, and T.~Lengauer, ``{Comparison of
  Classifier Fusion Methods for Predicting Response to Anti HIV-1 Therapy},''
  \emph{PLoS ONE}, vol.~3, no.~10, p. e3470, 2008.

\bibitem{Liu2012}
M.~Liu, D.~Zhang, and D.~Shen, ``{Ensemble Sparse Classification of Alzheimer's
  Disease},'' \emph{Neuroimage}, vol.~60, no.~2, pp. 1106--1116, 2012.

\bibitem{Khan2010}
A.~Khan, A.~Majid, and T.-S. Choi, ``{Predicting Protein Subcellular Location:
  Exploiting Amino Acid Based Sequence of Feature Spaces and Fusion of Diverse
  Classifiers},'' \emph{Amino Acids}, vol.~38, no.~1, pp. 347--350, 2010.

\bibitem{Pandey2010}
G.~Pandey, B.~Zhang, A.~N. Chang, C.~L. Myers, J.~Zhu, V.~Kumar, and E.~E.
  Schadt, ``{An Integrative Multi-Network and Multi-Classifier Approach to
  Predict Genetic Interactions},'' \emph{PLoS Computational Biology}, vol.~6,
  no.~9, p. e1000928, 2010.

\bibitem{Tumer1996}
K.~Tumer and J.~Ghosh, ``{Error Correlation and Error Reduction in Ensemble
  Classifiers},'' \emph{Connection Science}, vol.~8, no. 3-4, pp. 385--404,
  1996.

\bibitem{Kuncheva2003}
L.~I. Kuncheva and C.~J. Whitaker, ``{Measures of Diversity in Classifier
  Ensembles and Their Relationship with the Ensemble Accuracy},'' \emph{Machine
  Learning}, vol.~51, no.~2, pp. 181--207, 2003.

\bibitem{Dietterich2000}
T.~G. Dietterich, ``{An Experimental Comparison of Three Methods for
  Constructing Ensembles of Decision Trees: Bagging, Boosting, and
  Randomization},'' \emph{Machine Learning}, vol.~40, no.~2, pp. 139--157,
  2000.

\bibitem{Tang2006}
E.~K. Tang, P.~N. Suganthan, and X.~Yao, ``{An Analysis of Diversity
  Measures},'' \emph{Machine Learning}, vol.~65, no.~1, pp. 247--271, 2006.

\bibitem{Breiman1996}
L.~Breiman, ``{Bagging Predictors},'' \emph{Machine Learning}, vol.~24, no.~2,
  pp. 123--140, 1996.

\bibitem{Schapire2012}
R.~E. Schapire and Y.~Freund, \emph{{Boosting: Foundations and
  Algorithms}}.\hskip 1em plus 0.5em minus 0.4em\relax MIT Press, 2012.

\bibitem{Merz1999}
C.~J. Merz, ``{Using Correspondence Analysis to Combine Classifiers},''
  \emph{Machine Learning}, vol.~36, no. 1-2, pp. 33--58, 1999.

\bibitem{Wolpert1992}
D.~H. Wolpert, ``{Stacked Generalization},'' \emph{Neural Networks}, vol.~5,
  no.~2, pp. 241--259, 1992.

\bibitem{Caruana2004}
R.~Caruana, A.~Niculescu-Mizil, G.~Crew, and A.~Ksikes, ``{Ensemble Selection
  from Libraries of Models},'' in \emph{Proceedings of the 21st International
  Conference on Machine Learning}, 2004, pp. 18--26.

\bibitem{Caruana2006}
R.~Caruana, A.~Munson, and A.~Niculescu-Mizil, ``{Getting the Most Out of
  Ensemble Selection},'' in \emph{Proceedings of the 6th International
  Conference on Data Mining}, 2006, pp. 828--833.

\bibitem{Niculescu-Mizil2009}
A.~Niculescu-Mizil, C.~Perlich, G.~Swirszcz, V.~Sindhwani, and Y.~Liu,
  ``{Winning the KDD Cup Orange Challenge with Ensemble Selection},''
  \emph{Journal of Machine Learning Research Proceedings Track}, vol.~7, pp.
  23--24, 2009.

\bibitem{Pandey2006}
G.~Pandey, V.~Kumar, and M.~Steinbach, ``{Computational Approaches for Protein
  Function Prediction: A Survey},'' University of Minnesota, Tech. Rep., 2006.

\bibitem{Hess2009}
D.~C. Hess, C.~L. Myers, C.~Huttenhower, M.~A. Hibbs, A.~P. Hayes, J.~Paw,
  J.~J. Clore, R.~M. Mendoza, B.~S. Luis, C.~Nislow, G.~Giaever, M.~Costanzo,
  O.~G. Troyanskaya, and A.~A. Caudy, ``{Computationally Driven, Quantitative
  Experiments Discover Genes Required for Mitochondrial Biogenesis},''
  \emph{PLoS Genetics}, vol.~5, no.~3, p. e1000407, 2009.

\bibitem{Szczurek2013}
E.~Szczurek, N.~Misra, and M.~Vingron, ``{Synthetic Sickness or Lethality
  Points at Candidate Combination Therapy Targets in Glioblastoma},''
  \emph{International Journal of Cancer}, 2013.

\bibitem{Bella2012}
A.~Bella, C.~Ferri, J.~Hern\'{a}ndez-Orallo, and M.~J. Ram\'{\i}rez-Quintana,
  ``{On the Effect of Calibration in Classifier Combination},'' \emph{Applied
  Intelligence}, pp. 1--20, 2012.

\bibitem{Hughes2000}
T.~R. Hughes, M.~J. Marton, A.~R. Jones, C.~J. Roberts, R.~Stoughton, C.~D.
  Armour, H.~A. Bennett, E.~Coffey, H.~Dai, Y.~D. He, M.~J. Kidd, A.~M. King,
  M.~R. Meyer, D.~Slade, P.~Y. Lum, S.~B. Stepaniants, D.~D. Shoemaker,
  D.~Gachotte, K.~Chakraburtty, J.~Simon, M.~Bard, and S.~H. Friend,
  ``{Functional Discovery via a Compendium of Expression Profiles},''
  \emph{Cell}, vol. 102, no.~1, pp. 109--126, 2000.

\bibitem{Myers2006}
C.~L. Myers, D.~R. Barrett, M.~A. Hibbs, C.~Huttenhower, and O.~G. Troyanskaya,
  ``{Finding Function: Evaluation Methods for Functional Genomic Data},''
  \emph{BMC Genomics}, vol.~7, no. 187, 2006.

\bibitem{Pandey2009}
G.~Pandey, C.~L. Myers, and V.~Kumar, ``{Incorporating Functional
  Inter-Relationships Into Protein Function Prediction Algorithms},'' \emph{BMC
  Bioinformatics}, vol.~10, no. 142, 2009.

\bibitem{Hartman2001}
J.~L. Hartman, B.~Garvik, and L.~Hartwell, ``{Principles for the Buffering of
  Genetic Variation},'' \emph{Science}, vol. 291, no. 5506, pp. 1001--1004,
  2001.

\bibitem{Horn2011}
T.~Horn, T.~Sandmann, B.~Fischer, E.~Axelsson, W.~Huber, and M.~Boutros,
  ``{Mapping of Signaling Networks Through Synthetic Genetic Interaction
  Analysis by RNAi},'' \emph{Nature Methods}, vol.~8, no.~4, pp. 341--346,
  2011.

\bibitem{Vandersluis2010}
B.~VanderSluis, J.~Bellay, G.~Musso, M.~Costanzo, B.~Papp, F.~J. Vizeacoumar,
  A.~Baryshnikova, B.~J. Andrews, C.~Boone, and C.~L. Myers, ``{Genetic
  Interactions Reveal the Evolutionary Trajectories of Duplicate Genes},''
  \emph{Molecular Systems Biology}, vol.~6, no.~1, 2010.

\bibitem{R}
{R Core Team}, ``{R: A Language and Environment for Statistical Computing},''
  2012.

\bibitem{Hornik2009}
K.~Hornik, C.~Buchta, and A.~Zeileis, ``{Open-Source Machine Learning: R Meets
  Weka},'' \emph{Computational Statistics}, vol.~24, no.~2, pp. 225--232, 2009.

\bibitem{Hall2009}
M.~Hall, E.~Frank, G.~Holmes, B.~Pfahringer, P.~Reutemann, and I.~H. Witten,
  ``{The WEKA Data Mining Software: An Update},'' \emph{SIGKDD Explorations},
  vol.~11, no.~1, pp. 10--18, 2009.

\bibitem{Kuhn2008}
M.~Kuhn, ``{Building Predictive Models in R Using the caret Package},''
  \emph{Journal of Statistical Software}, vol.~28, no.~5, pp. 1--26, 2008.

\bibitem{Hothorn2010}
T.~Hothorn, P.~B\"{u}hlmann, T.~Kneib, M.~Schmid, and B.~Hofner, ``{Model-Based
  Boosting 2.0},'' \emph{Journal of Machine Learning Research}, vol.~11, no.
  Aug, pp. 2109--2113, 2010.

\bibitem{Friedman2010}
J.~H. Friedman, T.~Hastie, and R.~J. Tibshirani, ``{Regularization Paths for
  Generalized Linear Models via Coordinate Descent},'' \emph{Journal of
  Statistical Software}, vol.~33, no.~1, pp. 1--22, 2010.

\bibitem{Venables2002}
W.~N. Venables and B.~D. Ripley, \emph{{Modern Applied Statistics with S}},
  4th~ed.\hskip 1em plus 0.5em minus 0.4em\relax Springer, 2002.

\bibitem{Ridgeway2013}
G.~Ridgeway, ``{gbm: Generalized Boosted Regression Models},'' 2013.

\bibitem{Liaw2002}
A.~Liaw and M.~Wiener, ``{Classification and Regression by randomForest},''
  \emph{R News}, vol.~2, no.~3, pp. 18--22, 2002.

\bibitem{Hastie2002}
T.~Hastie, R.~J. Tibshirani, B.~Narasimhan, and G.~Chu, ``{Diagnosis of
  Multiple Cancer Types by Shrunken Centroids of Gene Expression},''
  \emph{Proceedings of the National Academy of Sciences of the United States of
  America}, vol.~99, no.~10, pp. 6567--6572, 2002.

\bibitem{Ting1999}
K.~M. Ting and I.~H. Witten, ``{Issues in Stacked Generalization},''
  \emph{Journal of Artificial Intelligence Research}, vol.~10, no.~1, pp.
  271--289, 1999.

\bibitem{Yule1900}
G.~U. Yule, ``{On the Association of Attributes in Statistics: With
  Illustrations from the Material of the Childhood Society},''
  \emph{Philosophical Transactions of the Royal Society of London, Series A},
  vol. 194, pp. 257--319, 1900.

\bibitem{Cohen1960}
J.~Cohen, ``{A Coefficient of Agreement for Nominal Scales},''
  \emph{Educational and Psychological Measurement}, vol.~20, no.~1, pp. 37--46,
  1960.

\bibitem{Eddelbuettel2011}
D.~Eddelbuettel and R.~Fran\c{c}ois, ``{Rcpp: Seamless R and C++
  Integration},'' \emph{Journal of Statistical Software}, vol.~40, no.~8, pp.
  1--18, 2011.

\bibitem{Demsar2006}
J.~Dem\v{s}ar, ``{Statistical Comparisons of Classifiers over Multiple Data
  Sets},'' \emph{Journal of Machine Learning Research}, vol.~7, no. Jan, pp.
  1--30, 2006.

\bibitem{Friedman1937}
M.~Friedman, ``{The Use of Ranks to Avoid the Assumption of Normality Implicit
  in the Analysis of Variance},'' \emph{Journal of the American Statistical
  Association}, vol.~32, no. 200, pp. 675--701, 1937.

\bibitem{Nemenyi1963}
P.~B. Nemenyi, ``{Distribution-Free Multiple Comparisons},'' Ph.D.
  dissertation, Princeton University, 1963.

\bibitem{Brown2010}
G.~Brown and L.~I. Kuncheva, ``{``Good" and ``Bad" Diversity in Majority Vote
  Ensembles},'' in \emph{Proceedings of the 9th International Conference on
  Multiple Classifier Systems}, 2010, pp. 124--133.

\bibitem{Li2012}
N.~Li, Y.~Yu, and Z.-H. Zhou, ``{Diversity Regularized Ensemble Pruning},'' in
  \emph{Proceedings of the 2012 European Conference on Machine Learning and
  Knowledge Discovery in Databases}, 2012, pp. 330--345.

\bibitem{Brier1950}
G.~W. Brier, ``\BIBforeignlanguage{EN}{{Verification of Forecasts Expressed in
  Terms of Probability}},'' \emph{\BIBforeignlanguage{EN}{Monthly Weather
  Review}}, vol.~78, no.~1, pp. 1--3, 1950.

\bibitem{Platt1999}
J.~C. Platt, ``{Probabilistic Outputs for Support Vector Machines and
  Comparisons to Regularized Likelihood Methods},'' \emph{Advances in Large
  Margin Classifiers}, vol.~10, no.~3, pp. 61--74, 1999.

\bibitem{Zadrozny2002}
B.~Zadrozny and C.~Elkan, ``{Transforming Classifier Scores Into Accurate
  Multiclass Probability Estimates},'' in \emph{Proceedings of the 8th ACM
  International Conference on Knowledge Discovery and Data Mining}, 2002, pp.
  694--699.

\end{thebibliography}

\end{document}